%% file: main-2595-Sabo.tex
\documentclass[11pt,a4paper]{article}
\usepackage{times,latexsym}
\usepackage{url}
\usepackage[T1]{fontenc}

\usepackage[acceptedWithA]{tacl2018v2}
\usepackage[]{tacl2018v2}

\usepackage{times}
\usepackage{graphicx}
\usepackage{multicol}
\usepackage{multirow}
\usepackage{booktabs}
\usepackage{tikz}
\usepackage{pgfplots}
\usepackage{adjustbox} 
\usepackage{lipsum}

\usepackage{latexsym}
\usepackage{outlines}
\usepackage{soul}
\usepackage{amsfonts}
\usepackage{graphicx}
\usepackage{subfig}
\usepackage{multicol}
\usepackage{microtype}
\usepackage{textcomp}
\usepackage{amssymb}

\usepackage{mathtools}
\usepackage{dsfont}
\usepackage{amsmath}

\usepackage{xspace,mfirstuc,tabulary}

\newif\iftaclinstructions
\taclinstructionsfalse 
\iftaclinstructions
\renewcommand{\confidential}{}
\renewcommand{\anonsubtext}{(No author info supplied here, for consistency with
TACL-submission anonymization requirements)}
\newcommand{\instr}
\fi

\iftaclpubformat 

\else

\fi

\newcommand{\osa}[1]{{#1}}
\newcommand{\os}[1]{{#1}}
\newcommand{\osnew}[1]{{#1}}

\newcommand{\RR}[1]{}
\newcommand{\ol}[1]{}

\newcommand{\nwks}{N-Way K-Shot}
\newcommand{\MTBEM}{BERT\textsubscript{EM}}
\newcommand{\sups}{Support Set}

\newcommand{\fstac}{Few-Shot TACRED}
\newcommand{\fs}{Few-Shot}

\title{Revisiting Few-shot Relation Classification: \\
Evaluation Data and Classification Schemes}

\author{Ofer Sabo\textsuperscript{1} \,
 Yanai Elazar\textsuperscript{1,2} \,
 Yoav Goldberg\textsuperscript{1,2} \,
 Ido Dagan\textsuperscript{1} \\
\textsuperscript{1}Computer Science Department, Bar Ilan University \\
\textsuperscript{2}Allen Institute for Artificial Intelligence \\
\\
  {\tt  \{ofersabo,yanaiela,yoav.goldberg,ido.k.dagan\}@gmail.com}
}

\date{}

\begin{document}
\setlength{\abovedisplayskip}{7pt}
\setlength{\belowdisplayskip}{9pt}
\maketitle
\begin{abstract}
  \input{Abstract}
\end{abstract}

\input{1_introduction.tex}

\input{2.0_task_formulation.tex}

\input{3.0_Desired_properties_and_existing}

\input{3.3_New_Few_Shot_Datasets_and_Methodology}

\input{4.0_Existing_Models}

\input{5.0_analysis_and_improvement.tex}

\input{5.3_MNAV}

\input{5.4_training_procedure}

\input{6.0_Experiments.tex}

\input{6.1_FewRel_experiments.tex}

\input{6.2_TACRED_experiments.tex}

\input{7.0_analysis.tex}

\input{8.0_conclusions_and_future_work.tex}

\bibliography{tacl2018}
\bibliographystyle{acl_natbib}

\newpage
\input{9.0_appendix}

\end{document}

%% file: Abstract.tex

\osnew{We explore Few-Shot Learning (FSL) for Relation Classification (RC).}
Focusing on the realistic scenario of FSL, in which a test instance might not belong to any of the target categories (none-of-the-above, aka NOTA), we first revisit the recent popular dataset structure for FSL, \osnew{pointing out} its unrealistic data distribution. To remedy this, we propose a novel methodology for deriving more realistic few-shot test data from available datasets for supervised RC, and apply it to the TACRED dataset. This yields a new challenging benchmark for FSL RC, on which state of the art models show poor performance. 
Next, we analyze classification schemes within the popular embedding-based nearest-neighbor approach for FSL, with respect to constraints they impose on the embedding space. Triggered by this analysis we propose a novel classification scheme, in which the NOTA category is represented as \osnew{learned} vectors, shown empirically to be an appealing option for FSL.

%% file: 1_introduction.tex
\section{Introduction}
\label{sec:introduction}

We consider relation classification (RC)---an important sub-task of relation extraction (RE)---in which one is interested in determining, giving a text with two marked entities, if the entities conform to one of pre-determined relations, or not.
While supervised methods for this task exist and work relatively well \cite{MTB,DBLP:conf/emnlp/Zhang0M18,DBLP:conf/acl/WangCML16,DBLP:conf/acl/MiwaB16}, they require large amounts of training data\osnew{,} which is hard to obtain in practice.

We are therefore interested in a data-lean scenario in which users provide only a handful of training examples for each relation they are interested in. This has been formalized in the ML community as \osnew{\fs{} Learning} (FSL) (\S\ref{TASK_SETUP}).

\begin{figure}[!t]
\centering
\includegraphics[width=7cm]{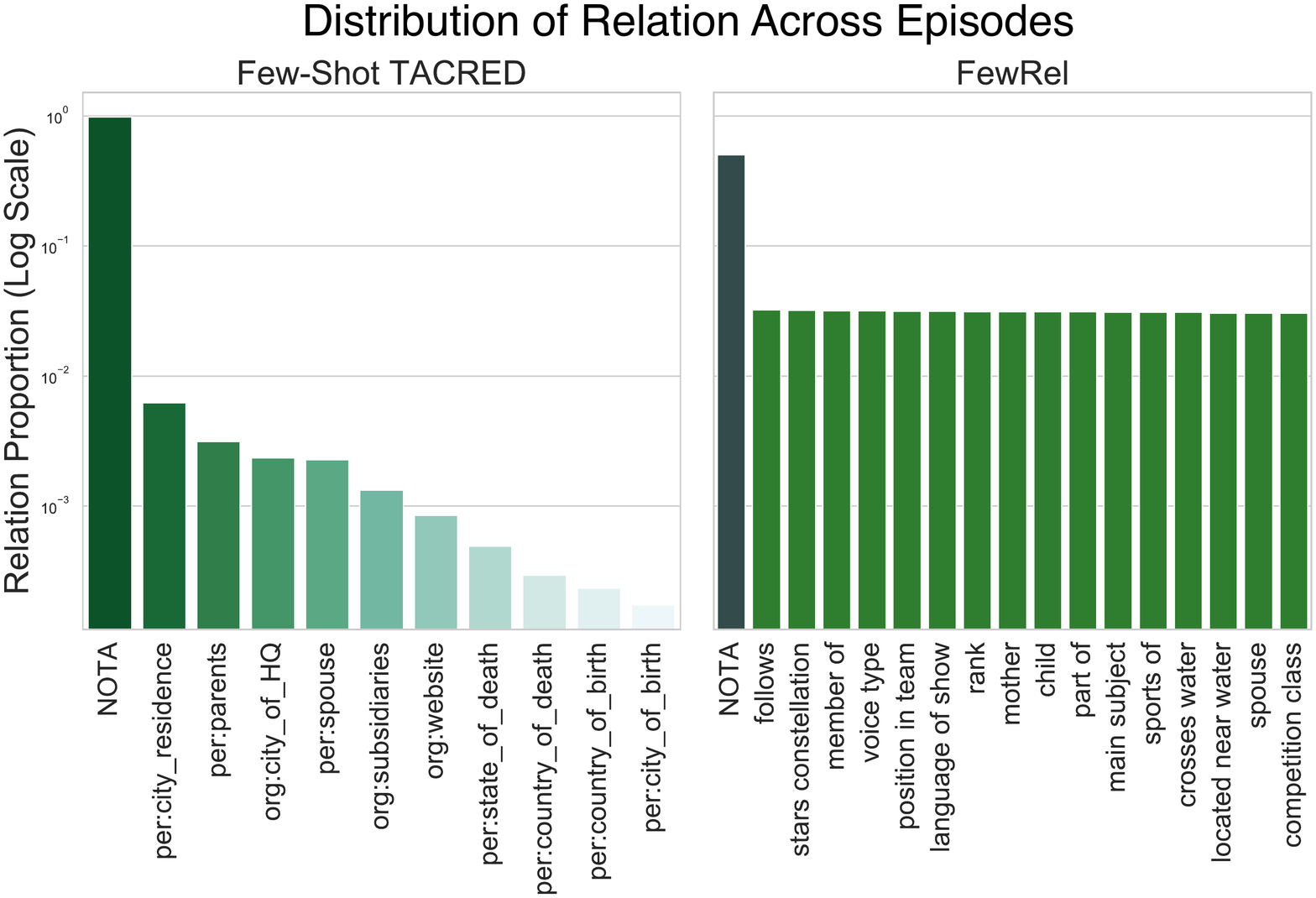}
\caption{Relation distribution across episodes in our newly derived \fstac{} and the existing FewRel 2.0 RC task. On the left side we demonstrate the relations distribution in \fstac{} episodes, which follows a real-world distribution.
On the right, we present the relations distribution in FewRel 2.0, which is synthetic. The y-axis for both figures is in log scale. \fstac{} NOTA's proportion is 97.5\% while in FewRel 2.0 it is 50\%.} 
\label{fig:Episodes_stats}
\end{figure}

FSL for Relation Classification has been recently addressed by the work of \citet{FEWREL,FEWREL_second}, who introduced the FewRel 1.0 and shortly after the FewRel 2.0 challenges, in which researchers are provided with a large labeled dataset of background relations, and are tasked with producing strong few-shot classifiers: classifiers that will work well given a few labeled examples of relations not seen in the training set. The task became popular, with scores on FewRel 1.0 achieving an accuracy of 93.9\% \cite{MTB}, surpassing the human level \os{performance} of 92.2\%. Results on FewRel 2.0 are lower, at 80.3\% for the best system \cite{FEWREL_second}), but are still very high considering the \osnew{difficulty of the task}.

Is few-shot relation classification solved? We show that this is far from being the case. We argue that the evaluation protocol in FewRel 1.0 is based on highly unrealistic assumptions on how the models will be used in practice, and while FewRel 2.0 tried to amend it, its evaluation setup remains highly unrealistic (\S\ref{sec:existing_fsl}). 
Therefore, we propose a methodology to transform supervised datasets into corresponding realistic few-shot evaluation scenarios (\S\ref{TRANSFORMATION}) .
We then apply our transformation on the supervised TACRED dataset \cite{tacredzhang2017} to create such a new few-shot evaluation set (\S\ref{sec:fstac_transformation}). Our experiments (\S\ref{sec:fstac_results}) reveal that indeed, moving to this realistic setup, the performance of existing State-Of-The-Art (SOTA) models drop considerably, from scores of around 80 F1 (as well as accuracy) to around 30.

A core factor in a realistic few-shot setup is the NOTA (none-of-the-above) option; allowing a case where a particular test instance does not conform to \emph{any} of the predefined target relations. Triggered by presenting an analysis of possible decision rules for handling the NOTA category (\S\ref{sec:few_shot_formulation}), we propose a novel  enhancement which models NOTA by an explicit set of \os{vectors} in the embedding space (\S\ref{sec:nota_as_vec}). 
This explicit ``NOTA as vectors'' approach achieves new SOTA performance for the FewRel 2.0 dataset, and outperforms other models on our new  dataset  (\S\ref{EVALUATION}).
Yet, the realistic scenario of our TACRED-derived dataset remains far from being solved, calling for substantial future research. We release our models, data, and, more importantly, our data conversion procedure, to encourage such future work.

%% file: 2.0_task_formulation.tex
\section{Task Setup and Formulation}
\label{TASK_SETUP}

\subsection{Relation Classification}
The \emph{relation extraction} (RE) task takes as input a set of documents and a list of
pre-specified relations, and aims to extract tuples of the form 
$(e_1, e_2, r)$ 
where $e_1$ and $e_2$ are entities, $r$ is a relation that holds between them ($r$ belongs to a pre-specified list of relations of interest). 
This task is often approached by a
pipeline that generates candidate $(e_1, e_2, s)$ triplets, classifies each one to a
relation (or indicates there is no relation).
The classification task from such triplets to an expressed relation is called
\emph{relation classification} (RC). It is often isolated and addressed on its
own, and is also the focus of the current work. \citet{tacredzhang2017} demonstrate that
improvements in RC carry over to improvements in RE.

In the RC task each input $x_i=(e_1, e_2, s)_i$ consists of a sentence $s$ with a (ordered) pair of \textit{marked
entities} (each entity is a span over $s$), and the output is one of $|R|+1$
classes, indicating that the entities in $s$ conform to one of the relations in a set 
$R$ of \emph{target relations}, or
to none of them. We refer to a triplet $x_i$ as a \emph{relation instance}.
For example, if the target relations are $R=\{$Owns, WorksFor$\}$,
the relation instance 
\textit{"Wired reports that in a surprising reshuffle at  \underline{Microsoft}$_{e_2}$, \underline{Satya Nadella}$_{e_1}$ has taken over as
the managing director of the company."} should be classified as WorksFor. The
same sentence with the entity pair $e_1=$\emph{Satya Nadella} and
$e_2=$\emph{Wired} should be classified as ``NoRelation'' (NOTA).


\subsection{The Few-Shot N-Way K-Shot Setup}
As supervised datasets are often hard and expensive to obtain, there is a growing interest in the
\emph{few-shot} scenario, where the user is interested in $|R|$ 
target-relations, but can provide only a few labeled instances for each relation.
In this work, we follow the increasingly popular \emph{N-Way K-Shot} setup of \osnew{Few-Shot} Learning (FSL), proposed by \citet{vinyals2016matching,snell2017prototypical}.
This setup was adapted to relation classification, resulting in the FewRel and FewRel 2.0 datasets \cite{FEWREL,FEWREL_second}. We further discuss the datasets in \S\ref{Desired}.

The N-Way K-Shot setup assumes the user is interested in $N$ target relations ($R^{target}=\{c_1,...,c_N\}$), and has access to $K$ instances (typically few) of each one, called the \emph{support set} for class $c_j$, denoted by $\sigma$:
\[
\sigma = \{\sigma_{c_1},...,\sigma_{c_N} \} \;\;\;\; c_{j}\in R^{target}
\]
\[
\sigma_{c_j} = \{x_1,...,x_k\} \;\;\; \text{s.t.} \; r(x_i) = c_j
\]
where $r(x)$ is the gold relation of instance $x$;
$\sigma_{c_j}$ is the support set for relation $c_j$; and
$\sigma$ is the support set for all $N$ relations in
$R^{target}$.

A set of target relations and the corresponding support sets is called a scenario. 
Given a scenario $\mathcal{S}=(R^{target},\sigma)$, our goal is to create a decision function $f_\mathcal{S}(x): x \to R^{target}\cup \{\perp \}$, where $\perp$ indicates ``none of the relations in $R^{target}$''. 
Let $X=x_1,...,x_m$ be a set of instances with corresponding true labels $r(x_1),...,r(x_m)$, our aim is to minimize the average cumulative evaluation loss $\frac{1}{m}\sum_{i=1}^{m}\ell(f_S(x_i),r(x_i))$, where $\ell$ is a per-instance loss function, usually zero-one loss.

When treating FSL as a transfer learning problem, as we do here, there is also a background set of relations $R^{background}$, disjoint from the target relation set, for which there is plenty of labeled data available. This data can also be used for constructing the decision function.

The performance of an \nwks{} FSL algorithm on a dataset $X$ is highly dependent on the specific scenario $\mathcal{S}$: both the choice of the the $N$ relations that needs to be distinguished 
as well as the choice of the specific K examples for each relation 
can greatly influence the results. 
In a real-life scenario, the user is interested in a specific set of relations and examples, but when developing and evaluating FSL algorithms, we are concerned with the \emph{expected performance} of a method given an arbitrary set of categories and examples:
$\mathbb{E}_\mathcal{S}[\frac{1}{m}\sum_{i=1}^{m}\ell(f_S(x_i),r(x_i))]$ which can be approximated by averaging the losses for several random scenarios $\mathcal{S}_j$, each varying the relation set and the example set. In a practical evaluation, the number of \nwks{} scenarios that can be considered is limited, relative to the combinatorial number of possible scenarios. To maximize the number of considered scenarios, we re-write the loss to consider expectations also over the data points: $\mathbb{E}_\mathcal{S}\mathbb{E}_{(x)\sim X}[\ell(f_\mathcal{S}(x),r(x))]$. 

This gives rise to an evaluation protocol that considers the loss over many \emph{episodes}, where each episode is composed of: (1) a random choice $R^{target}$ of $N$ distinct target relations $R^{target}=\{c_1,...,c_N\},\; c_i \neq c_j$; (2) a corresponding random support set $\sigma=\{\sigma_{c_1},...,\sigma_{c_N}\}$ of $N * K$ instances ($K$ instances in each $\sigma_{c_j}$); and (3) a \emph{single} randomly-chosen labeled example considered as  a \emph{query}, $(x,r(x))$, which does not appear in the support set. 
To summarize, an evaluation set for \nwks{} FSL is a set of episodes, each consisting of a $N$ target relations, $K$ supporting examples for each relation, and a query. For each episode, the algorithm should classify the query instance to one of the relations in the support set, or none of them.

In practice, the episodes in an evaluation set are obtained by sampling episodes from a labeled dataset. As we discuss in the following section, the specifics of the labeled dataset and the sampling procedure can greatly influence the realism of the evaluation, and the \osnew{difficulty} of the task. 

\subsection{\osnew{Low-resource Relation Classification --- Related Work}}
\osnew{Other than FSL, several setups for investigating RC under low resource setting have been proposed.}

\osnew{\citet{DBLP:conf/acl/ObamuyideV19} experimented with limited supervision settings on TACRED. Their setting is different though than the transfer-based few-shot setting, addressed in our paper. In most of their experiments the amount of training instances per relations is much higher, not fitting the ad-hoc nature of the few-shot setting.
Further, they train a model on all classes, not addressing inference on new class types at test time.}


\osnew{Distant supervision is another approach for handling low-resource RC \cite{DBLP:conf/acl/MintzBSJ09}. This approach leverages noisy labels for training a model, produced by aligning relation instances to a knowledge-base. Particularly, it considers sentences containing a pair of entities holding a known relation as instances of that  relation. For example, a sentence containing the entities `Barack Obama', and  `Hawaii' will be labeled as an instance of the \textit{born\_in} relation between these entities, even though that sentence might describe, for example, a later visit of Obama to Hawaii.}

\osnew{Finally, another line of work is the Zero-Shot setup, where the RC task is reduced to another inference task, leveraging trained models for that task. Specifically,  \citet{levy-etal-2017-zero} proposed a method that leverages reading comprehension models, while \citet{obamuyide-vlachos-2018-zero} suggest using textual entailment models.}

%% file: 3.0_Desired_properties_and_existing.tex
\section{Desired Versus Existing Few-Shot Relation Classification Datasets}
\label{Desired}
A FSL system is intended to be used in a real-life scenario. Thus, evaluation procedures for FSL should attempt to mimic the conditions under which the FSL system will be applied in practice. In a realistic FSL scenario, the user has a set of relations of interest (``target relations"), and can come up with a handful of examples for each. The relations in the set are often related to each other. 
The user may potentially have access to a \emph{labeled dataset} of a \emph{different set} of relations (``background relations"), which they may want to use to train, or to improve, their FSL system. 

The resulting classifier will then be applied to unlabeled data aiming to detect new \textit{target} relations, in which, realistically: 

\noindent\textbf{(a)} some relations are rarer than others.

\noindent\textbf{(b)} most instances do not correspond to a target relation.

\noindent\textbf{(c)} many instances may not correspond also to a background relation.

\RR{Reviewer 2: last point}

\os{\noindent\textbf{(d)} relation instances may include named entities, as well as pronouns and common noun entities.
}

Ideally, the episodes in an FSL evaluation should be chosen in a way that reflects (a)-(\os{d}) above.\footnote{ Additional concern of a realistic setup, which we do not consider in this work, is the accuracy of the entity-extractor that marks entity boundaries and assigns entity types, prior to the RC setup.} \RR{Reviewer 2: last point}
\os{The first characteristic (a) naturally follows the non-uniform distribution of relation types in a (non-artificial) text collection. The second point (b) stems from the fact that a natural text refers to a broad, inherently unbound, range of relation types, while in an RC setting, particularly for FSL, there is typically a restricted set of target relations. Similarly, while available RC training sets (for the supervised setting) may annotate more relation types than in a typical few-shot setting, they still contain a limited number of relation types in comparison to the full range of relations expressed in the corpus. 
This is prototypically evidenced in the naturally distributed  RC dataset TACRED (\S \ref{sec:fstac_transformation}), where 78.56\% of the labels are NOTA (Table \ref{tab:TACRED_single_spilt}).
Finally, naturally occurring textual relations may be used to relate named entities as well as common nouns or pronouns (d); therefore, we expect the annotated RC dataset entities to include all such entity types.}

As we show below, existing FSL-RC datasets do not conform to these properties, resulting in artificial---and substantially easier---classification tasks. This in turn leads to inflated accuracy numbers that are not reflective of the real potential performance of a system. We propose a refined sampling procedure that adheres to the realistic setting, and results in a substantially more realistic evaluation set, while conforming to the same \nwks{} protocol. As we show in the experiments section (\S\ref{EVALUATION}), this setup proves to be substantially more challenging for existing algorithms. We propose to use this procedure for future evaluation of FSL-RC algorithms, and release the corresponding code and data.\footnote{Code repository to transform TACRED into \fstac{} \url{https://github.com/ofersabo/Few_Shot_transformation_and_sampling}.}

\subsection{Existing FSL RC Datasets}
\label{sec:existing_fsl}
An \nwks{} RC dataset was introduced by \citet{FEWREL}, called FewRel 1.0. The dataset became popular, yet proved to be rather easy: the current best leaderboard entry by \citet{MTB} obtain results of over 93.86\% accuracy for 5-way 1-shot, above the 92\% accuracy of human performance. The dataset was then updated to FewRel 2.0 \cite{FEWREL_second}, using an updated episode sampling procedure (see below), with the current best system obtaining a 5-way 1-shot score of 80.31 \cite{FEWREL_second}.

\paragraph{Underlying labeled data} Both FewRel versions are based on the same underlying labeled dataset containing 100 distinct relations, with 700 instances per relation, totalling in $70,000$ labeled instances. The sentences are based on Wikipedia and the entities and relation labels are assigned automatically using Wikidata, followed by a human verification step.

Note that while extensive, each relation type contains the same number of instances, regardless \osnew{of any real} truthful distribution in a corpus, resulting in a highly synthetic dataset, contradicting the realistic assumption (a) above. In contrast, instances in supervised RC datasets such as TACRED and DocRED \cite{tacredzhang2017,DocRED} do respect the relation distribution in a real corpus.

\RR{Reviewer 2: last point}
Finally, FewRel target entities are mostly named entities, not including important entity types such as pronouns and common nouns, which are present in supervised RC datasets (including TACRED), thus contradicting assumption (d).

\paragraph{Train/Dev/Test splits} The 100 relations are split into three disjoint sets, $R_{train}$, $R_{dev}$, and $R_{test}$, consisting of 64, 16 and 20 relations, respectively. The relations in $R_{train}$ and their corresponding instances are used as the labeled corpus of background relations, while evaluation episodes consist of relations in either $R_{dev}$ or $R_{test}$. We refer to this set (either test or dev) as $R_{eval}$. Each episode consists of random subset $R^{target} \subset R_{eval}$. 

\paragraph{Sampling procedures}
The episode sampling procedure of FewRel 1.0 works by sampling $N$ relations from $R_{eval}$ resulting in a target set $R^{target}$, sampling a corresponding size $k$ support set $\sigma_{c_j}$ for each $c_j\in R^{target}$, and then sampling a query example in which $r(q) \in R^{target}$. That is, the query in each episode is guaranteed to be in $R^{target}$. This setup is artificial, negating realistic condition (b) above. This explains the high performance on FewRel 1.0.

\paragraph{NOTA} Following the aforementioned observation, the FewRel 2.0 work introduced 
a none-of-the-above (NOTA) scenario. Here, after sampling the target relation set $R^{target} \subset R_{eval}$, the query class $r$ is sampled from $R^{target}$ with probability $p$ and from $R_{eval}\setminus R^{target}$ with probability $1-p$. That is, $1-p$ of the episodes contain a query for which the answer does not correspond to any support set, in which case the answer is NOTA.

While a step in the right direction (indeed, results in this setup drop from over 90\% to around 80\%), this setup is still highly unrealistic: not only all the NOTA instances are guaranteed to be valid relations, they also always come from the same small set, contradicting assumption (c). In a realistic setup, we would expect the vast majority of test instances to be NOTA, but the set of NOTA instances is expected to vary greatly: some of them will correspond to relations from the background relations, some of them will correspond to unseen relations, and many will not correspond to any concrete relation. Furthermore, some of the NOTA cases will appear in sentences that do contain a target relation, but between different entities. Supervised relation extraction and relation classification datasets reflect this situation, and we argue that the FSL evaluation sets should also do so.

%% file: 3.3_New_Few_Shot_Datasets_and_Methodology.tex
\subsection{Better FSL-RC Evaluation Sets} 
\label{TRANSFORMATION}

We propose a methodology for transforming a supervised RC dataset into a \osnew{few-shot} RC dataset, while attempting to maintain properties (a)-(\os{d}) of the realistic evaluation scenario.
\os{This methodology can be applied to existing and future supervised datasets, thus reducing the need of collecting new dedicated FSL datasets.}

\subsubsection{Realistic underlying labeled data}
We assume a given supervised dataset, with $C$ categories, divided into train and test sections, where each section contains all $C$ categories, with distinct instances in each section (the typical setting for supervised multi-class classification). 
Some instances (in all sections) may be labeled with ``None-of-the-above`` (also known as ``other`` in the classic supervised setting, or ``no relation'' in TACRED terminology), hereafter \textit{NOTA}, meaning these instances do not belong to any of the $C$ categories.

\paragraph{Transformation} 
We transform the supervised dataset into an FSL dataset containing (as in FewRel) a set of background relations for training and a disjoint set of relations for evaluation. 
To perform this transformation, we begin by choosing $M$ categories as $R_{eval}$.\footnote{In practice we have  $M_T$ categories for test and $M_D$ for dev, we refer to both as eval for brevity.}  The remaining $C - M$ categories are designated as background relations $R_{train}$.\footnote{To preform the \nwks~setup, $M$ is required to be larger than $N$; in case the original data has a NOTA label, $M$ may be equal to $N$.}
We now keep the same instance-level train/dev/test splits of the original supervised dataset, but relabel the instances in each section: train set instances whose labels are in $R_{train}$ retain their original labels, while all other training instances are labeled as NOTA. Similarly for the test and dev splits. This results in sets where each set has distinct labels, but some of the NOTA instances in one set correspond to labels in other sets. 

\paragraph{Multiple splits} 
The choice of relations for each set influences the resulting dataset: some relations are more similar to each other than others, and splits that put several similar relations in an eval set are harder than splits in which similar relations are split between the train an eval sets. Moreover, as the number of labeled instances for each relation differ, splitting by relation results in different number of train/dev/test instances. We thus repeat the process several times, each time with a distinct set of eval relations.

\subsubsection{Realistic episode sampling}

To create an episode, we first sample the $N*K$ instances, which constitute the $N$ \osnew{support set} as in previous episodic sampling: sample $N$ out of $M$ relations, and then sample $K$ instances for each relation from the underlying eval set. However, the query for the episode is then sampled uniformly from \emph{all} remaining instances in the eval set.  
If the label of the instance chosen as query differs from the $N$ target relations in the episode, it is labeled as NOTA.
This query sampling procedure maintains both the label distribution and NOTA rate of the underlying supervised dataset.

\subsection{\fstac{}: Realistic Few-Shot Relation Classification}
\label{sec:fstac_transformation}

\input{tables/Final_TACRED_split_single}
We apply our transformation methodology to the TACRED Relation Classification dataset \cite{tacredzhang2017}. \RR{Reviewer 2: point 1}\os{The TACRED dataset was collected from a news corpus, purposing extracting relations involving 100 target entities. Accordingly, each sentence containing a mention of one of these target entities was used to generate candidate relation instances for the RC task. 
The relation label was annotated as one of $41$ pre-defined relation categories, when appropriate, or into an additional ``no\_relation'' category. The ``no\_relation'' category corresponds to cases where some other relation type holds between the two arguments, as well as cases in which no relation holds between them, where we consider both types of cases as falling under our NOTA category.}

We choose $M=10$ of the $41$ relations for the test set, and divide the remaining 31 relations into 25 and 6 for training and development, respectively, and release this split for future research.
Table \ref{tab:TACRED_single_spilt} lists the respective number of train/dev/test instances in \os{our \fstac{}}, along with the resulting NOTA rate in the test instances, as well as the corresponding numbers for the original TACRED dataset. 
As we expected, in a typical \osnew{few-shot} setting over  natural text (as in \fstac{}, unlike FewRel), where the number of the targeted classes (N-way) is small, most instances would correspond to the NOTA case. This is indeed illustrated in Table \ref{tab:TACRED_single_spilt}, where the original TACRED dataset includes 41 target classes, vs. 10 in \fstac{},
and hence have a lower NOTA rate (conversely, in a 5-way setting, the NOTA rate is even higher, see Figure \ref{fig:numeros_NOTA_rates}). 

\paragraph{Evaluation sets}
For evaluation, we consider sets of 150,000 episodes, sampled according to the procedure above. For robustness, we create 5 evaluation sets of 30,000 episodes each, and report the mean and STD scores over the 5 sets. 
Figure \ref{fig:Episodes_stats} (shown in \S\ref{sec:introduction}) presents the distribution differences between \fstac{} and FewRel 2.0 episodes.  
As we show in Section \ref{EVALUATION}, the \fstac{} evaluation set proves to be a substantial challenge \osnew{for} Few-Shot algorithms.

%% file: tables/Final_TACRED_split_single.tex
\begin{table}[ht!]
\centering
\resizebox{0.9\columnwidth}{!}{%
\begin{tabular}{l|rrr}
Dataset & Train & Dev  & Test             \\ \hline
TACRED          & 13,012 & 5,436 & 3,325  (78.56\%) \\ \hline
FS TACRED & 8,163  & 633  & 804 (94.81\%)  \\
\end{tabular}%
}
\caption{\os{Number of relation instances in the original TACRED dataset and in our derived \fstac{}.The corresponding test set NOTA rates appear in parenthesis.}}
\label{tab:TACRED_single_spilt}
\vspace{-.3cm}
\end{table}

%% file: 4.0_Existing_Models.tex
\section{Background: Prior Few-Shot RC Models}
\label{sec:EXISTING_MODEL}

As mentioned earlier, only a handful of examples are provided for the target classes in the \osnew{Few-Shot} setting. 
It is therefore quite challenging to utilize these examples effectively for learning or updating model parameters. Consequently, quite a few existing \osnew{few-shot} models, in the machine learning literature as well as in NLP \cite{vinyals2016matching,ravi2016optimization,MTB},
perform a representation learning phase (typically known as \textit{embedding learning} \osnew{or \textit{metric learning}} ), followed by nearest neighbor classification.
Here, model parameters are first learned over the background classes, for which substantial training is available. Then, classification of test instances is based on the trained model, with the hope that this model would generalize reasonably well for the target classes.\footnote{While in the remainder of this paper we focus on this similarity-based approach, it is worth noting that there exist other approaches for \osnew{Few-Shot} Learning\osnew{,} which further utilize the few labeled support examples. These include \textit{data augmentation} methods, which generate additional examples based on the few initial ones, as well as \textit{optimization-based} methods  \cite{ravi2016optimization,maml_finn}, where the model does utilize the small support sets of the target classes for parameter learning. Integrating our contributions with these approaches is left for future work.}

In the nearest neighbor approach, classification is done via a scoring function $score(q,c_i)$, which assigns a score for a query instance, $q$, and a target class, $c_i$.
Since the class is represented by its \sups, $\sigma_{c_i}$, the scoring function can be a similarity function between the query and the class's support set:
\begin{equation*}
    score(q,c_i) \triangleq sim(q,\sigma_{c_i})
\end{equation*}
Most often, an embedding-based approach is taken to compute similarity, decomposing the process into two separate components \cite{snell2017prototypical,MTB,li_vision}. 
First, instances are embedded into an explicit, typically dense, vector space, by an embedding function. Then, query-support similarity is measured over embedded vectors. 
Specifically, the prototypical network of  \citet{snell2017prototypical} represents a target class $c_i$ by a class prototype vector $\mu_i$, which is the average embedding of the $K$ instances in the support set of the class. The similarity between the query and each support set, $sim(q, \sigma_{c_i})$, is then measured as the similarity between the query and the corresponding prototype vector, assuming some similarity function between vectors in the embedding space:
\begin{equation*}
    sim(q,\sigma_{c_i})  \triangleq sim(q,\mu_i)
\end{equation*}
This approach was adopted in the state-of-the-art method \cite{MTB} for \osnew{few-shot} Relation Classification (FewRel 1.0, excluding the NOTA category), as well as by several other works for FSL in NLP \cite{DBLP:journals/corr/abs-1908-06039,prototypical_text_clasification}.

\paragraph{Nearest-neighbor classification rule}
Similarity is computed between a test instance and each support set, selecting the nearest class: 
\[
f_\mathcal{S}(q) = \arg\max_{c_j} sim(q, \sigma_{c_j})
\]

\paragraph{Instance representation}

\citet{MTB} further conducted an empirical analysis of embedding functions for \osnew{few-shot} Relation Classification. Their most effective embedding method augments the sentence with special tokens at the beginning and at the end of each of the two entities of the relation instance.  The  instance representation is then obtained by concatenating the two corresponding \textit{start} tokens from BERT's last layer \cite{BERT}. 
In our experiments, we \os{adopt} this embedding function, denoted \MTBEM{} (BERT-based Entity Marking), as well as the use of dot product as the vector similarity function (after we reassessed its effectiveness as well).

\subsection{FewRel 2.0 BERT sentence-pair model}
\label{subsec:sentence-pair}\RR{Reviewer 2: point 3 and Reviewer 3 last point} 
The FewRel 2.0 work presented a \os{model} for the NOTA setting, which skips the embedding learning phase \cite{FEWREL_second}. Instead, it utilizes the embeddings-based next sentence prediction score of BERT  \cite{BERT}, as the similarity score between a query and each support set instance.
Then, similarly to the approach described above, a nearest-neighbor criterion is applied over the average similarity score between the query and all support instances of each class.
A parallel scoring mechanism is implemented to decide whether the NOTA category should be chosen.

\osnew{\subsection{Related FSL Classification Models}
In this section we first review some prominent Few-Shot Learning work addressing other machine-learning tasks. Additionally, we compare between the notions of Out-Of-Domain detection and NOTA detection.}

\osnew{In a recent work on FSL, \citet{Tseng2020Cross-Domain} aim to improve generalization abilities by providing supervision for the category transfer phase. In their learning setting, the classes of each training episode are divided into two subsets, the first acts as the ``typical" training set while the second simulates the test set. To improve generalization they add an additional encoding layer which is optimized to maximaize performance on the simulated test categories.}   


\osnew{Another recent FSL work, addressing text classification, suggests to weigh words by their frequency over the training set \cite{Bao2020Few-shot}. 
The model employs two components to classify the given text into one of the episode's categories.
The first component computes the inverse frequency of each support set token over the training set. The second component estimates the inductive level of support set tokens with respect to classification. Finally, the output of these two components is used to train a linear classifier, by which the query is classified.}

\osnew{\paragraph{Out-Of-Domain detection} The essence of the NOTA category resembles Out-Of-Domain detection, as in both cases the goal is to detect instances not falling under the known categories. \citet{DBLP:conf/emnlp/TanYWWPCY19} define the OOD classes as the set of all classes which were not part of the training classes (vs. NOTA, which means that none of the given support classes in an episode is present). }
\osnew{In their work, the authors suggest a representation learning approach for Out-Of-Domain (OOD) detection in text classification. Their method combines hinge loss with the classic cross-entropy loss function. The former is used to push away the representation of the OOD instances, while the latter is used to learn correct classification within the in-domain classes.}

%% file: 5.0_analysis_and_improvement.tex
\section{Classification Rules: Analysis and Extension}
\label{sec:few_shot_formulation}

In this section, we provide an analytic perspective on the bias \osnew{that} different nearest-neighbor 
classification rules impose on the learned embedding space. 
We start with an analysis of the classification rule for the basic \osnew{few-shot} RC setting, without the NOTA category, as was applied in prior work (Section \ref{sec:EXISTING_MODEL}). This analysis follows directly the constraint presented in the influential work of \citet{DBLP:journals/jmlr/WeinbergerS09} \RR{Review 3: point 1}, and utilized in subsequent work (e.g. \cite{DBLP:journals/corr/abs-1003-0487,DBLP:conf/acl/DhillonTC10}).
We then extend this analysis to the setting which does include the NOTA category. First, we analyze the straightforward threshold-based approach for this setting. 
Then, inspired by this analysis, we propose an alternative approach, with a corresponding constraint, which represents the NOTA category by one or more explicit learned vectors. 
As shown in subsequent sections, this new approach performs consistently better than other methods on both the FewRel 2.0 and our new \fstac{} benchmarks, and is thus suggested as an appealing approach for \osnew{few-shot} Relation Classification. 

\subsection{Constraints Imposed by Nearest-neighbor Classification}

\paragraph{Classification without NOTA}
As described earlier, the nearest neighbor approach assigns a query instance to the class of its nearest support set. 
We start our analysis by adapting inequality (10) in \citet{DBLP:journals/jmlr/WeinbergerS09}, which was introduced to formulate the training goal for metric learning in \textit{k}-nearest neighbor classification.
To \osnew{this} end, we adapt the original inequality to our nearest-neighbor few-shot classification setting (Section \ref{sec:EXISTING_MODEL}). The obtained inequality below specifies the\osnew{, necessary and sufficient constraints} that the embedding space, along with the similarity function over it, should satisfy in order to reach \textit{perfect} classification, over all possible episodes in a given dataset.\footnote{Notice that we drop the margin element in the adapted inequality, as it is not needed for the analytic purpose of our constraint.} 
For every possible query instance $q$, a support set $\sigma_{r(q)}$ from the same class as $q$ and a support set $\sigma_{\neg r(q)}$ for a different class, the following constraint should hold:
\begin{equation}
\begin{split}
\label{objective_without_NOTA}
\forall\; &q,\sigma_{r(q)},\sigma_{\neg r(q)} \;\\
& sim(q,\sigma_{r(q)}) > sim(q,\sigma_{\neg r(q)})
\end{split}
\end{equation}
That is, to achieve perfect classification, each possible relation instance $q$ imposes that support sets of different classes should be positioned further away from it (being less similar) than the most distant support set it might have from its own class. Generally speaking, the nearest neighbor classification rule implies that instances that are rather close to their class mates may also be rather close to other classes, while instances that are far from their class mates should also be positioned at least as far from other classes.

In the \osnew{few-shot} setting, the embedding function is learned during training, over the training categories. As the learning process tries to optimize classification on the training set, it effectively attempts to learn an embedding function that would satisfy the above constraint as much as possible. 
Indeed, we often observed almost perfect performance over the training data, indicating that, for the training instances, this constraint is mostly satisfied by the learned embedding function. 
Yet, while it is hoped that the embedding function would separate properly also instances of new, previously unseen, classes, in practice this holds to a lesser degree, as indicated by lower test \osnew{performance}.

\paragraph{Thresholded classification with NOTA}
When the NOTA option is present, the nearest neighbor classification rule can be naturally augmented by assigning the NOTA category to test queries whose similarity to all of the target classes does not surpass a predetermined (possibly learned) threshold, $\theta$.
Extending our analysis to such classification rule, to achieve \textit{perfect} classification, the embedding space must fulfil the following\osnew{, necessary and sufficient} constraint, whose left-hand-side is relevant only for episodes that include a support set for the query's class:
\begin{equation}
\label{objective_with_threshold}
\begin{split}
\forall\; q,&\sigma_{r(q)},\sigma_{\neg r(q)} \\
 & sim(q,\sigma_{r(q)}) > \theta > sim(q,\sigma_{\neg r(q)})
\end{split}
\end{equation}
Since the same threshold is applied to all queries, to achieve \textit{perfect} classification in this setting $\theta$ should be smaller than all within-class similarities, for any possible pair of query $q$ and a support set of its class $\sigma_{r(q)}$. Concurrently, it should be larger than all cross-class similarities, for any possible query $q$ and a support set of a different class $\sigma_{\neg r(q)}$.\footnote{\osnew{Proof provided in the appendix.}}

We observe that Inequality \ref{objective_with_threshold} imposes a \textit{global} constraint over the embedding space.
It implies that the degree to which all classes should be separated from each other is imposed, globally, by those queries in the entire space which are the furthest away from their own class support sets. Accordingly, it requires \textit{all} classes to be positioned equally far from each other, regardless of their own ``compactness". This makes a much harsher constraint, and challenge for the embedding learning, than Inequality (\ref{objective_without_NOTA}), which allows certain classes to be nearer if their within-class similarities are high.

\subsection{NOTA As a Vector (NAV)}
\label{sec:nota_as_vec}
Motivated by the last observation, we propose an alternative classification approach for \osnew{few-shot} classification with the NOTA category. In this approach, we represent the NOTA category by an \textit{explicit} vector in the embedding space, denoted $V_N$, which is learned during training. At test time, the similarity between the query $q$ and this vector, $sim(q,V_N)$, is computed and regarded as the similarity between the query and the NOTA category:
\begin{equation*}
    sim(q,\text{NOTA})  \triangleq sim(q,V_N)
\end{equation*}
Then, $q$ is assigned to its nearest class, by the usual nearest-neighbor classification rule. Thus, the NOTA class is selected if $sim(q,V_N)$ is higher than $q$'s similarity to all target classes. 
Effectively, this mechanism considers an \textit{individual} NOTA classification threshold for each query, namely $sim(q,V_N)$, which depends on $q$'s position in the embedding space relative to $V_N$. 
We term this approach \textit{``NOTA As a Vector``} (NAV).

Classification under the NAV scheme implies the following constraint on the embedding space, considering perfect classification:\footnote{Notice the analogous structure of Inequalities \ref{objective_with_threshold} and \ref{objective_with_nav}, 
where $sim(q,V_N)$ replaces the role of $\theta$. A similar correctness proof applies.}
\begin{equation}
\label{objective_with_nav}
\begin{split}
\forall \; &q,\sigma_{r(q)},\sigma_{\neg r(q)} \\
& sim(q,\sigma_{r(q)}) > sim(q,V_N) > sim(q,\sigma_{\neg r(q)}) 
\end{split}
\end{equation}

This constraint implies that, to achieve perfect classification, the similarity between a query and the NOTA vector $V_N$ should be smaller than $q$'s similarity to all possible support sets of its own class, while being larger than its similarity to all support sets of other classes. In comparison to the prior classification rules, this approach does allow instances that are rather close to their class mates to be closer to other classes than instances that are positioned further from some of their class mates, similarly to the lighter constraint in Inequality (\ref{objective_without_NOTA}). Yet, to enable such ``geometry" of the embedding space, it is also required that instances would be positioned appropriately relative to the NOTA vector, in a way that satisfies the two constraints in \osnew{Inequality} (\ref{objective_with_nav}). Using the NAV approach, it is hoped that the learning process would position the NOTA vector, and adjust the embedding parameters, such that these constraints would be mostly satisfied. 
Overall, the NAV approach imposes different constraints on the similarity space than using a single global classification threshold for the NOTA category (as in \osnew{Inequality} (\ref{objective_with_threshold})), and it is not clear apriori which approach would be more effective to learn. This question is investigated empirically in Section \ref{EVALUATION}.

%% file: 5.3_MNAV.tex
\subsection{Multiple NOTA vectors}
\label{sec:ׁMNAV}

A natural extension of the NAV approach, denoted as \textit{MNAV}, is to represent the NOTA category by \textit{multiple} vectors, whose number is an empirically tuned hyper-parameter.
During classification, the model picks the closest vector to the query as $V_N$, which accordingly defines $sim(q,\text{NOTA})$. Then, classification is determined as in the NAV method,
where adding multiple NOTA vectors is expected to effectively ease the embedding space constraints.
\osnew{In practice, we treat the number of NOTA vectors as a hyperparameter.}

%% file: 5.4_training_procedure.tex
\subsection{Training Procedure}

For training, we use the same episode sampling procedure that generated the dev/test sets, but where the target relations are sampled from a set of train relations, disjoint from the dev/test relations. We define an epoch to include a fixed number of episodes, considered a tuned hyper\osnew{-}parameter, independently sampling episodes for each epoch. We measure dev set performance after each epoch, and use early stopping.
For each episode $\mathcal{E}=(R^{target},\{\sigma_{c_1},...,\sigma{c_N}\},q)$, we encode the query using \MTBEM{} encoding function \cite{MTB}, described in \S\ref{sec:EXISTING_MODEL}, $\vec{q}=\text{\MTBEM}(q)$ and similarly for each item $x$ in each support set, obtaining for each $\sigma_{c_j}$ the corresponding average prototype vector $\vec\mu_j = \frac{1}{K}\sum_{x\in{\sigma_{c_j}}}\text{\MTBEM}(x)$. 

We define the prototype of the NOTA class to be the learned NAV vector: $\vec\mu_{\perp}=\vec{v}_N$. Our loss term for each episode considers $\vec{q}$ and the prototype vectors $\vec\mu_i$ and tries to optimize \osnew{Inequality} (3): $dot(\vec{q},\vec\mu_{r(q)})>dot(\vec{q},\vec\mu_{\perp})>dot(\vec{q},\sigma_{\neg r(q)})$.
Concretely we use \emph{cross-entropy loss}, as used in previous work \cite{MTB}:
\[
-\log\frac{e^{dot(\vec{q},\vec\mu_{r(q)})}}
{\sum_{i \in R^{target}\cup\{\perp\}}e^{dot(\vec{q},\vec{\mu_i})}}
\]
Note that this works towards satisfying the conditions in \osnew{Inequality} (3): in episodes where $r(q)\neq\perp$, the loss attempts to increase the first term in \osnew{Inequality} (3) (the similarity between the query and the prototypical vector of its class), while decreasing the similarity of the two other terms (the similarity between $q$ and all other prototypical vectors, including the NAV one). In particular, it drives towards satisfying  $sim(q,\sigma_{r(q)}) > sim(q,V_N)$. In episodes where $r(q)=\perp$, the loss increases the second term, decreasing the similarity in the third term, driving towards satisfying 
$sim(q,V_N)>sim(q,\sigma_{\neg r(q)})$.
Analogously, the same dynamics apply when the learned (scalar) threshold value determines the NOTA score.

Following \citet{DBLP:journals/jmlr/WeinbergerS09}, who derived a triplet loss objective, and similar to subsequent lines of work (e.g. \cite{DBLP:journals/corr/SchroffKP15,DBLP:journals/corr/HofferA14,DBLP:conf/iccvw/MingCLVB17}), we experimented also with adapted versions of triplet loss. 
Under this objective, instances not belonging to the same class are pushed away while same-class instances are pulled together, aiming to reach the desired ordering as in inequalities \osnew{\ref{objective_with_threshold} and \ref{objective_with_nav}}. We tried multiple variants of this objective for FSL training, including objective versions with a margin element, but these experiments resulted in consistently lower results than the methods described above.

\paragraph{NOTA vectors initialization}
For the NAV method, we straightforwardly initialized the single NOTA vector randomly. 
Random initialization of the multiple NOTA vectors in MNAV evolved to a single vector being dominantly picked as the NAV vector by the MNAV decision process. Consequently, results were very similar to the (single vector) NAV model.
Presumably, this happened because a single random vector turned out to be closest to the sub-space initially populated by the pre-trained \MTBEM{} embedding function.
To avoid this, we wish to scatter all the initial vectors within the initially populated subspace. 
\osnew{To this end, we initialize a NOTA vector by sampling a relation and then averaging 10 random instances from that relation. We repeat this process for each NOTA vector.}

%% file: 6.0_Experiments.tex
\section{Experiments and Results}
\label{EVALUATION}
In this section\osnew{,} we assess our two main contributions. With respect to our \fstac{} dataset, we show that models that perform well on FewRel 2.0 perform poorly on this much more realistic setting, leaving a huge gap for improvement by future research. With respect to our proposed \osnew{NOTA As Vectors} modeling approach, we show that it is a viable, and advantageous, alternative to the threshold approach.

\paragraph{Implemented models}
We conduct our investigation in the framework of the common embedding based approach to FSL, with respect to the MNAV, NAV, and threshold-based methods described in \S\ref{sec:few_shot_formulation}. 
These methods are implemented following the \osnew{best-performing} embedding and similarity methods identified for the \osnew{state-of-the-art} method on FewRel 1.0 \cite{MTB}, namely \MTBEM{} applied using BERT\textsubscript{BASE}, and dot product similarity (\S\ref{sec:EXISTING_MODEL}). 
\osnew{In addition, we train and evaluate the baseline Sentence-Pair model, described in \S\ref{subsec:sentence-pair}.}

\osnew{
To select the number of NOTA vectors in the MNAV model, we experimented with 5 different values, ranging from 1 to 20. In practice, the choice of the number of vectors had rather little impact on the results (less than one F1 point). We use the best performing value for this hyperparameter, which was 20.}

\osnew{In terms of memory utilization, as 5-way 5-shot episodes require feeding the 25 instances of the support set in addition to the query instances into BERT simultaneously, they often occupy nearly the entire 32GB of GPU memory. To leverage the memory taken by the support set instances, we include as many queries as we can fit into the GPU's memory. In our experiments, we construct 3 episodes for each sampled support set (by sampling 3 different queries for it), which fully utilizes the GPU capacity. 
Since these episodes occupy the entire GPU memory, we use a single episode per batch. 
}

\osa{We further note that it may be possible to perform the N-way classification by transforming it into a pair-wise classification, repeated N times (both in training and evaluation). This technique would allow to reduce the memory usage but would increases the run-time.
As we managed to fit the entire episode to our GPU memory, we followed the standard N-way approach, for faster computation, as was previously done by \citet{FEWREL_second}.}

\paragraph{Test methodology and metrics}
Like prior work, evaluation is conducted over randomly sampled episodes from the test data, as described in \S\ref{TASK_SETUP}.
Prior results for FewRel 2.0 (and FewRel 1.0) were reported in terms of Accuracy. However, in realistic, \osnew{highly imbalanced}, relation classification datasets, like our \fstac{}, accuracy becomes meaningless.
Hence, we propose micro F1 over the target relations as a more appropriate measure for future research. Accordingly, we report micro F1 for both datasets, as well as accuracy for FewRel experiments, for compatibility. For both measures we report average values and standard deviation over 5 different random samples of episodes \cite{DBLP:conf/emnlp/Zhang0M18,tacredzhang2017}. In all experiments, we train and evaluate five models and report the results of the median performing model. \RR{Reviewer 1: point 1 and Reviewer 3: last point} Unless otherwise mentioned, reported result differences are significant under one-tailed t-test at 0.05 confidence.

%% file: 6.1_FewRel_experiments.tex
\subsection{FewRel 2.0 Result}
\label{sec:fewrel_results}

\input{tables/Final_FewRel_test_set}

We first confirm the appropriateness of our investigation by comparing  performance on the prior FewRel 2.0 test data. Table \ref{tab:FewRel_test_set} presents the figures on the two official (synthetic) test NOTA rates for this benchmark.
We use 50\% NOTA rate to train all our models, with 6,000 episodes per epoch. As shown, the MNAV model performs best across all FewRel settings, obtaining a new SOTA for this task.\footnote{Our MNAV results are also reported at the official FewRel 2.0 leader-board, as \textit{Anonymous Cat}, at \url{https://thunlp.github.io/2/fewrel2_nota.html}. We note that the FewRel test set is kept hidden, where models are submitted to the FewRel authors, who produce (only) accuracy scores.}

\input{tables/Final_fewrel_dev_set_with_mnav}

We next turn to a more comprehensive comparison of the investigated embedding-based few-shot models, namely threshold-based, NAV, and MNAV, over the publicly available FewRel development set, with 50\% NOTA rate. The results in Table \ref{tab:Fewrel_dev_set_with_mnav} show that, 

here as well, the MNAV model outperforms the others in both settings. The gap between MNAV and the threshold model is significant for the two settings, while the gap relative to the NAV model is significant only in the 5-shot setting.

%% file: tables/Final_FewRel_test_set.tex
\begin{table}[t]
\centering
\resizebox{\columnwidth}{40pt}{
\begin{tabular}{l|cl|ll}
\hline
Model & \multicolumn{2}{c|}{5-way 1-shot} & \multicolumn{2}{l}{5-way  5-shot} \\
NOTA Rate & \multicolumn{1}{l}{15\%} & 50\% & 15\% & 50\% \\ \hline \hline
Sentence-Pair & 77.67 & \multicolumn{1}{c|}{80.31} & 84.19 & 86.06 \\
Threshold & \multicolumn{1}{l}{63.41} & 76.48 & 65.43 & 78.95 \\
NAV & 77.17 & \multicolumn{1}{c|}{81.47} & 82.97 & 87.08 \\
MNAV & \textbf{79.06} & \multicolumn{1}{c|}{\textbf{81.69}} & \textbf{85.52} & \textbf{87.74} \\ \hline
\end{tabular}%
}
\caption{Accuracy results on FewRel2.0 test set, for the four available settings for this benchmark. Results are reported for the FewRel2.0  sentence-pair baseline model and our investigated  models.}
\label{tab:FewRel_test_set}
\vspace{-.2cm}
\end{table}

%

%% file: tables/Final_fewrel_dev_set_with_mnav.tex
\begin{table}[t!]
\centering
\scalebox{0.75}{
\begin{tabular}{llcc}
\hline
Model                      & Metric   & 5-way 1-shot                           & 5-way 5-shot                           \\ \hline
\multirow{2}{*}{Sentence-Pair} & Accuracy & $75.48\pm{0.33\%}$                     & $78.43\pm{0.25\%}$                     \\ \cline{2-4} 
& F1       & \multicolumn{1}{l}{$71.85\pm{0.44\%}$} & \multicolumn{1}{l}{$75.43\pm{0.31\%}$} \\ \hline
\multirow{2}{*}{Threshold} & Accuracy & $76.32\pm{0.12\%}$                     & $80.30\pm{0.09\%}$                     \\ \cline{2-4} 
                           & F1       & \multicolumn{1}{l}{$73.34\pm{0.25\%}$} & \multicolumn{1}{l}{$78.89\pm{0.11\%}$} \\ \hline
\multirow{2}{*}{NAV}       & Accuracy & $78.54\pm{0.08\%}$                     & $80.44\pm{0.11\%}$                     \\ \cline{2-4} 
                           & F1       & \multicolumn{1}{l}{$75.00\pm{0.22\%}$} & \multicolumn{1}{l}{$79.20\pm{0.14\%}$} \\ \hline
\multirow{2}{*}{MNAV}      & Accuracy & $78.23\pm{0.13\%}$                     & $81.25\pm{0.18\%}$                     \\ \cline{2-4} 
 & F1 & \multicolumn{1}{l}{$\mathbf{75.22\pm{0.19\%}}$} & \multicolumn{1}{l}{$\mathbf{80.06\pm{0.11\%}}$} \\ \hline
\end{tabular}%
}
\caption{FewRel2.0 development set results, accuracy and micro F1.}
\label{tab:Fewrel_dev_set_with_mnav}
\end{table}

%% file: 6.2_TACRED_experiments.tex
\subsection{\fstac{} Results}
\label{sec:fstac_results}

\input{tables/Final_TACRED_test_set_3_only_micro_f1}

We compare the MNAV\os{, NAV, Sentence-Pair} and threshold-based models over our more realistic \fstac{} test set (here, epoch size is 2000).
As seen in Table \ref{tab:TACRED_3_split_results}, the MNAV model outperforms the others, as was the case over FewRel 2.0.

Notably, performance is drastically lower over \fstac{}. We suggest that this indicates the much more challenging nature of a realistic setting, relative to the FewRel 2.0 setting, while indicating the limitation of all current models. We further analyze this performance gap in the next section.

%% file: tables/Final_TACRED_test_set_3_only_micro_f1.tex
\begin{table}[t]
\centering
\scalebox{0.75}{
\begin{tabular}{c|cc}
\hline
model     & 5-way 1-shot             & 5-way 5-shot             \\ \hline
Sentence-Pair & $10.19\pm{0.81\%}$ & $-$ \\ \hline
Threshold & $6.87\pm{0.48\%}$ & $13.57\pm{0.46\%}$ \\ \hline
NAV       & $8.38\pm{0.80\%}$ & $18.38\pm{2.01\%}$ \\ \hline
MNAV      & $\mathbf{12.39}\pm{\mathbf{1.01\%}}$ & $\mathbf{30.04}\pm{\mathbf{1.92\%}}$ \\ \hline
\end{tabular}%
}
\caption{\os{Micro F1 results on \fstac{}. \osnew{For computational memory limitations, we could not evaluate the Sentence-Pair model in the 5-shot setting, see Appendix for explanation.}
}}
\vspace{-.3cm}
\label{tab:TACRED_3_split_results}
\end{table}

%% file: 7.0_analysis.tex
\section{Analysis}

\subsection{Differentiating characteristics of FewRel vs. \fstac{}}
As seen in Tables \ref{tab:Fewrel_dev_set_with_mnav} vs\@  \ref{tab:TACRED_3_split_results}, the results on \fstac{} are drastically lower than those obtained for FewRel 2.0, by at least 50 points. 
Yet, the performance figures are difficult to compare due to several differences between the datasets, including training size, NOTA rate, and different entity types. To analyze the possible impact of these differences, we control for each of them and observe performance differences. For brevity, we focus on the MNAV model (1-shot and 5-shot).

\paragraph{Training size}
We train the model on FewRel 2.0, taking the same amount of training instances as in \fstac{}. 
Compared to full training, results dropped by five micro F1 points in the 1-shot setting and by 1.5 points for 5-shot, suggesting that the training size explains only a small portion of the performance gap between the two datasets.

\paragraph{NOTA rates}

\begin{figure}[!t]
\centering
\includegraphics[width=1.0\columnwidth]{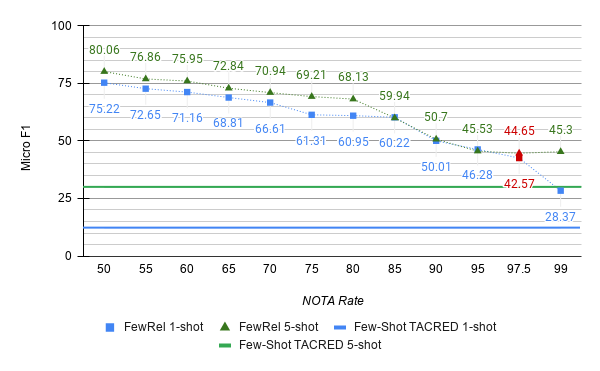}
\caption{MNAV results on the FewRel 2.0 dev dataset at different NOTA rates. The red points represent performances at 97.5\% NOTA rate which is the \fstac{} NOTA rate. \osnew{The blue and green horizontal lines denote the \fstac{} performance in the 1 and 5 shot settings, respectively.}}
\label{fig:numeros_NOTA_rates}
\end{figure}

We control for the unrealistic NOTA rate in FewRel 2.0 by training and evaluating our model on higher NOTA rates. The results in Figure \ref{fig:numeros_NOTA_rates} indicate that realistic higher NOTA rates are indeed much more challenging: moving from the original FewRel 50\% NOTA rate to the 97.5\% rate as in \fstac{} shrank the performance gap by 33 points in the 1-shot setting and by 35 for 5-shot.

\paragraph{Entity types}

In this experiment, we evaluate performance differences when including all entity types (named entities, common nouns and pronouns), as in \fstac{}, versus including only named entities, as in FewRel. \RR{Reviewer 2: last point } To this end, we sampled two corresponding subsets of relation instances from \fstac{}, of the same size, with either all entity types or named entities only.\footnote{Entity types were automatically identified by SpaCy NER model \cite{spacy2}, as well as certain fixed types included in FewRel, such as ranks and titles.} 
Further, we control for the distributions of relation types in the two subsets, making them equal, since, as discussed in Section \ref{Desired}, this distribution impacts performance in RC datasets. 

Apparently, the impact of entity composition was different in the 1-shot and 5-shot settings. 
For 1-shot, the named entities subset yielded slightly lower performance (6.65 vs. 9.03 micro F1), which is hard to interpret. For 5-shot, performance on the named entities subset was substantially higher than when including all entity types (33.48 vs. 18.74), possibly suggesting that a larger diversity of entity types is more challenging for the model. In any case, we argue that RC datasets should include all entity types, to reflect real-world corpora.

\paragraph{Summary}
Overall, the differences we analyzed account for much of the large performance gap between the two datasets, particularly in the more promising 5-shot setting. As argued earlier, we suggest that \fstac{} represents more realistic properties of few-shot RC, including realistic non-uniform distribution, ``no\_relation'' instances and inclusion of all entity types, and hence should be utilized in future evaluations.

\subsection{Few-Shot versus Supervised TACRED}
We next analyze the impact of category transfer in \fstac{}. 
To this end, we apply our same MNAV model in a supervised (non-transfer) setting, termed Supervised MNAV, and compare it to the few-shot MNAV (FSL MNAV).
Concretely, we trained the supervised MNAV model on the training instances of the \textit{same} categories as those in the \fstac{} test data (vs. training on different background relations in the transfer-based FSL setting).
The supervised model was then tested for 5-way 5-shot classification on \fstac{}, identically to the FSL MNAV 5-way 5-shot testing in Table \ref{tab:TACRED_3_split_results}. 
The results
showed a 31 point gap, with the Supervised MNAV yielding 61.19 micro F1 while FSL MNAV scored 30.04, indicating the substantial challenge when moving from the supervised to the category transfer setting.

\subsection{Qualitative Error Analysis}

To obtain some insight on current performance, we manually analyzed 50 episodes for which the model predicted an incorrect support class (precision error) and 50 in which it missed identifying the right support class (recall error). We sampled 1-shot episodes since these can be more easily interpreted, examining a single support instance per class.

For the precision errors, we found a single prominent characteristic. Across all sampled episodes, both the query and the falsely selected support instance shared the same (ordered) pair of entity types.
For instance, they may both share the entity types of \textit{person} and \textit{location}, albeit having different relations, such as \textit{city of death} vs. \textit{state of residence}, or having no meaningful relation for the query (\textit{no relation} case).
This behavior suggests that pre-training, together with fine tuning on the background relations, allowed the BERT-based model to learn to distinguish entity types, to realize their criticality for the RC task, and to successfully match entity types between a query and a support instance. On the other hand, the low overall performance suggests that the model does not recognize well the patterns indicating a target relation based on a small support set.
Additional evidence for this conjecture is obtained when examining confused class pairs in the predictions' confusion matrices (1-shot and 5-shot settings). 
Out of 10 confused class pairs, 8 pairs have matching entity types; in the other two pairs, the \textit{location} type is confused with \textit{organization} in the context of \textit{school attended}, which often carries a sense of location.

For the recall errors, manual inspection of the 50 episodes did not reveal any prominent insights. Therefore, we sampled $100,000$ 1-shot episodes over which we analyzed various statistics which may be related to recall errors. Of these, we present two analyses that seem to explain aspects of recall misses, in a statistically significant manner (one-tailed t-test at 0.01 significance level), though only to a partial extent. 

The first analysis examines the impact of whether the relative order of the two marked argument entities flips between the query and support instance sentences. To that end, we examined the about $2,600$ episodes in our sample in which the query belongs to one of the support classes. 
We found that for episodes in which argument order is consistent across the query and support instance, the model identified the correct class in  $15.68\%$ of the cases, while when the order is flipped only $10.95\%$ of the episodes are classified correctly. This suggests that a flipped order makes it more challenging for the model to match the relation patterns across the query and support sentences.
The second analysis examines the impact of lexical overlap between the query and support instance. To that end, we compared 300 episodes in which the correct support class was successfully identified (true positive) and 300 in which it was missed (false negative). In each episode, we measured Intersection over Union (IoU) (aka Jaccard Index) for the two sets of lemmas in the query and support instance. As expected, the IoU value was significantly higher for the true positive set (0.17) that for the false negative set (0.12), suggesting that higher lexical match eases recognizing the correct support instance.

%% file: 8.0_conclusions_and_future_work.tex
\section{Conclusions}

In this work, we lay several required criteria for realistic FSL datasets,  while proposing a methodology to derive such benchmarks from available datasets designed for supervised learning. We then applied our methodology on the TACRED relation classification dataset, creating a challenging benchmark for future research. Indeed, previous models that achieved impressive results on FewRel, a synthetic dataset for FSL, failed miserably on our naturally distributed dataset. These results call for better models and loss functions for FSL, and indicate that we are far from having satisfying results on this setup. Our methodology may be further applied to additional datasets, enriching the availability of realistic datasets for FSL. 

Next, we analyzed the constraints imposed embedding functions by nearest-neighbor classification schemes, common for FSL. This analysis led us to derive a new method for representing the NOTA category as one or more  explicit learned vectors, yielding a novel classification scheme, which achieves new state-of-the-art performance. We suggest that our analysis may further inspire additional innovations in \osnew{few-shot} learning.

%% file: 9.0_appendix.tex
\appendix

\section{Appendix}
\subsection{Proof of Inequality \ref{objective_with_threshold}}

We prove that the two sides of the inequality are necessary and sufficient to guarantee perfect classification by the threshold-based classification rule, over all possible episodes in a given dataset.

We first prove necessity. As the LHS refers to $\sigma_{r(q)}$, it is relevant only for episodes where $q$ belongs to one of the support classes. If it is violated for some episode, then that episode cannot be classified to $r(q)$ (the correct class) by the threshold-based classification rule.
As for RHS necessity, consider an episode in which 
$sim(q,\sigma_{\neg r(q)}) > \theta$, violating the RHS. 
Without loss of generality, we can construct a possible episode with the same $q$ and 
$\sigma_{\neg r(q)}$, whose correct classification is NOTA (making sure to exclude $r(q)$ from the support classes).
This episode cannot be correctly classified by the classification rule to NOTA, since $q$'s similarity to at least one class, $\neg r(q)$, surpasses $\theta$.

To prove sufficiency, we consider the two cases where an episode's correct classification is either NOTA or one of the support classes. 
If the correct classification is NOTA, then $r(q)$ is not within the \sups{}. The RHS then guarantees that $sim(q,\sigma_{\neg r(q)}) < \theta$ for all support classes, implying a correct NOTA classification.
Otherwise, the correct classification is $r(q)$, being one of the support classes.
In this case, the LHS guarantees excluding a NOTA classification, while the RHS excludes classification to any other category different than $r(q)$. QED.


\subsection{Sentence-Pair High GPU Demand }
\ol{
* why sentence-pair requires 4 times more memory
}

\osnew{The Sentence-Pair model \cite{FEWREL_second} requires at least twice more GPU memory than a standard embedding learning model, such as the threshold model (described in Sec \ref{sec:few_shot_formulation}). The higher memory demand arises from feeding BERT with the concatenation of each support instance to each query instance. This concatenation effectively doubles the average input sentence length. Due to the Transformer architecture, doubling the input sentence length requires higher GPU RAM memory. In particular, a fully connected layer requires four times more memory when fed with a double-length sequence. Hence, representation learning models, which embed a single instance into an embedded vector space, are more memory efficient than the sentence-pair model.}

\osnew{As mentioned in Section \ref{sec:fstac_results}, we could not train the sentence-pair model for the \fstac{} 5-shot setting, due to memory limitations, even though we used NVIDIA TESLA V100-32GB GPU. This stems from the fact that the average sentence length in  \fstac{} is higher than in FewRel, which did not fit into our server memory with  the higher memory consumption of the sentence-pair model.}

